\documentclass[a4paper,journal]{IEEEtran}

\usepackage{cite}
\usepackage{amsmath,amssymb,amsfonts}
\usepackage{algorithm,algpseudocode}
\usepackage{graphicx}
\usepackage{textcomp}
\usepackage{xcolor}
\usepackage{booktabs}
\usepackage{multirow}
\usepackage{tikz}
\usetikzlibrary{shapes,arrows,positioning,decorations.pathreplacing,fit,shapes.geometric}
\usepackage{pgfplots}
\pgfplotsset{compat=1.17}

\usepackage[caption=false,font=footnotesize]{subfig}

\algnewcommand\algorithmicinput{\textbf{Input:}}
\algnewcommand\INPUT{\item[\algorithmicinput]}
\algnewcommand\algorithmicoutput{\textbf{Output:}}
\algnewcommand\OUTPUT{\item[\algorithmicoutput]}
\algnewcommand{\algorithmicand}{\textbf{ and }}
\algnewcommand{\algorithmicor}{\textbf{ or }}
\algnewcommand{\OR}{\algorithmicor}
\algnewcommand{\AND}{\algorithmicand}

\makeatletter
\newcounter{phase}[algorithm]
\newlength{\phaserulewidth}
\newcommand{\setphaserulewidth}{\setlength{\phaserulewidth}}
\newcommand{\phase}[1]{%
	\vspace{-1.25ex}
	\Statex\leavevmode\llap{\rule{\dimexpr\labelwidth+\labelsep}{\phaserulewidth}}\rule{\linewidth}{\phaserulewidth}
	\Statex\strut\refstepcounter{phase}\textit{\textbf{Layer~\thephase~--~#1}}
	\vspace{-1.25ex}\Statex\leavevmode\llap{\rule{\dimexpr\labelwidth+\labelsep}{\phaserulewidth}}\rule{\linewidth}{\phaserulewidth}}
\makeatother
\setphaserulewidth{.7pt}

\def\BibTeX{{\rm B\kern-.05em{\sc i\kern-.025em b}\kern-.08em
		T\kern-.1667em\lower.7ex\hbox{E}\kern-.125emX}}

\begin{document}

\title{Hierarchical Federated Graph Attention Networks for Scalable and Resilient UAV Collision Avoidance}

\author{Rathin Chandra Shit and Sharmila Subudhi
	\thanks{Rathin Chandra Shit, Ph.D.  (e-mail: rathin088@gmail.com).}
	\thanks{Sharmila Subudhi, Ph.D. (e-mail: sharmilasubudhi@ieee.org).}
}

\maketitle

\begin{abstract}
The real-time performance, adversarial resiliency, and privacy preservation are the most important metrics that need to be balanced to practice collision avoidance in large-scale multi-UAV (Unmanned Aerial Vehicle) systems. Current frameworks tend to prescribe monolithic solutions that are not only prohibitively computationally complex with a scaling cost of $O(n^2)$ but simply do not offer Byzantine fault tolerance. The proposed hierarchical framework presented in this paper tries to eliminate such trade-offs by stratifying a three-layered architecture. We spread the intelligence into three layers: an immediate collision avoiding local layer running on dense graph attention with latency of $<10 ms$, a regional layer using sparse attention with $O(nk)$ computational complexity and asynchronous federated learning with coordinate-wise trimmed mean aggregation, and lastly, a global layer using a lightweight Hashgraph-inspired protocol. We have proposed an adaptive differential privacy mechanism, wherein the noise level $(\epsilon \in [0.1, 1.0])$ is dynamically reduced based on an evaluation of the measured real-time threat that in turn maximized the privacy-utility tradeoff. Through the use of Distributed Hash Table (DHT)-based lightweight audit logging instead of heavyweight blockchain consensus, the median cost of getting a $95^{th}$ percentile decision within 50ms is observed across all tested swarm sizes. This architecture provides a scalable scenario of 500 UAVs with a collision rate of $< 2.0\%$ and the Byzantine fault tolerance of $f < n/3$.
\end{abstract}


\section{Introduction} \label{sec1}
Large swarms of Unmanned Aerial Vehicle (UAV) in common airspace create highly demanding safety and security issues with tight real-time requirements. Under adversarial conditions, collision avoidance should work in sub-50ms (from here onward, ms be interpreted as milliseconds) decision windows with malicious nodes that might seek to jam the communication, impute false data or poison individual UAVs through a gradient poisoning attack \cite{chang2023distributed, talaat2024blockchain}. Although Federated Learning (FL) and Graph Attention Networks (GATs) have potential to be distributed coordination tools, a straightforward mixture of the two can produce computationally expensive $O(n^2)$ optimization procedures which can only cooperate with 50 UAVs \cite{zeng2025intersection, guo2025hybrid}.

ORCA (Optimal Reciprocal Collision Avoidance) \cite{alagha2025uav} and RVO2 (Reciprocal Velocity Obstacle) \cite{sunshine2019rvo} are classical implementation methods that paint geometric collision avoidance with collision avoidance certainties. However, they do not have the capacity of learning advanced scenarios. Recent deep reinforcement learning solution developed by Wang et al. \cite{wang2025blockchain} provides a better performance, but in need of a centralized coordination \cite{guillen2022multi, kwon2019multi}. Thus, it introduces bottlenecks and a single point of failure while avoiding collisions.

Further, Standard FL solutions have synchronization delay issues and are not Byzantine resilient \cite{chang2023distributed}. Talat and Hamza \cite{talaat2024blockchain} suggested an asynchronous FL work to enhance real-time throughput, but fails to deal with the peculiarity of collision avoidance where high latency can create issues \cite{lu2020blockchain}. In addition, the Graph Attention Networks (GATs) recently demonstrated the potential of graph based models in multi-agent coordination, although they exhibit quadratic complexity scaling. Sparse attention mechanisms \cite{guo2025hybrid} are less computationally demanding but they have not been systematically considered on the real-time UAV coordination with Byzantine fault tolerance \cite{li2021byzantine}.

These existing frameworks have a fundamental trilemma in terms of scalability-security guarantees. To the best of our knowledge, these issues are: 
\begin{itemize}
\item They either trade off real-time performance and security guarantees (blockchain-based solutions, which must accept $>500\,ms$ consensus to be safe and require broadcasting raw sensor data \cite{liu2020blockchain}), or
\item They scale with security guarantees (centralized approaches which have to be vulnerable to single point of failure), or
\item They apply blockchain consensus to every collision avoidance decision. 
\end{itemize}
The Monolithic architectures via application of blanket security to all the interactions are resource-wasting, and often too expensive \cite{wang2025blockchain}. Further, the dense graph attention is too fast to be possible on large swarms \cite{wang2025fedfr}.

\textbf{Key Contributions:}  This paper solves the scalability-security trilemma by stratifying the architecture. We suggest the structure in the form of the hierarchy of sources of resources to perform the computation and safeguard based on the situation within operation and time demands.
\begin{enumerate}
    \item A \textbf{three-layer hierarchical structure} to draw the decision of collision avoidance based on the latency requirements (0-10ms local, 10-30ms regional, 30-50ms global) through sparse temporal graph attention to attain $O(nk)$ complexity where $k << n$.
    \item A real-time \textbf{Asynchronous Federated Learning (AFL) protocol} with coordinated trimmed mean aggregation that yields $f < n/3$ Byzantine fault tolerance without significant performance overheads.
    \item An \textbf{adaptive differential privacy mechanism} that dynamically optimizes privacy budgets based on the quantified threat scores, to strike the privacy-utility trade-off at the most desired accuracy under the mixtures of threats. The noise levels $(\epsilon \in [0.1, 1.0])$ based on these quantified threat scores optimizes the privacy-utility trade-off up to 15\% accuracy improvement under low-threat conditions.
    \item An \textbf{Audit system based on lightweight DHT} with zero-knowledge proofs that significantly (85\%) decreases the overhead logging requirement associated with blockchain-based solutions and ensures cryptographic integrity.
\end{enumerate}
Our framework offers a realistic approach to launch UAV swarms of up to 500 agents in a manner that ensures safety guarantees and adversarial resilience.

\section{System Architecture} \label{sec2}
\begin{figure*}[!ht]
\centering
\includegraphics[width=0.6\textwidth]{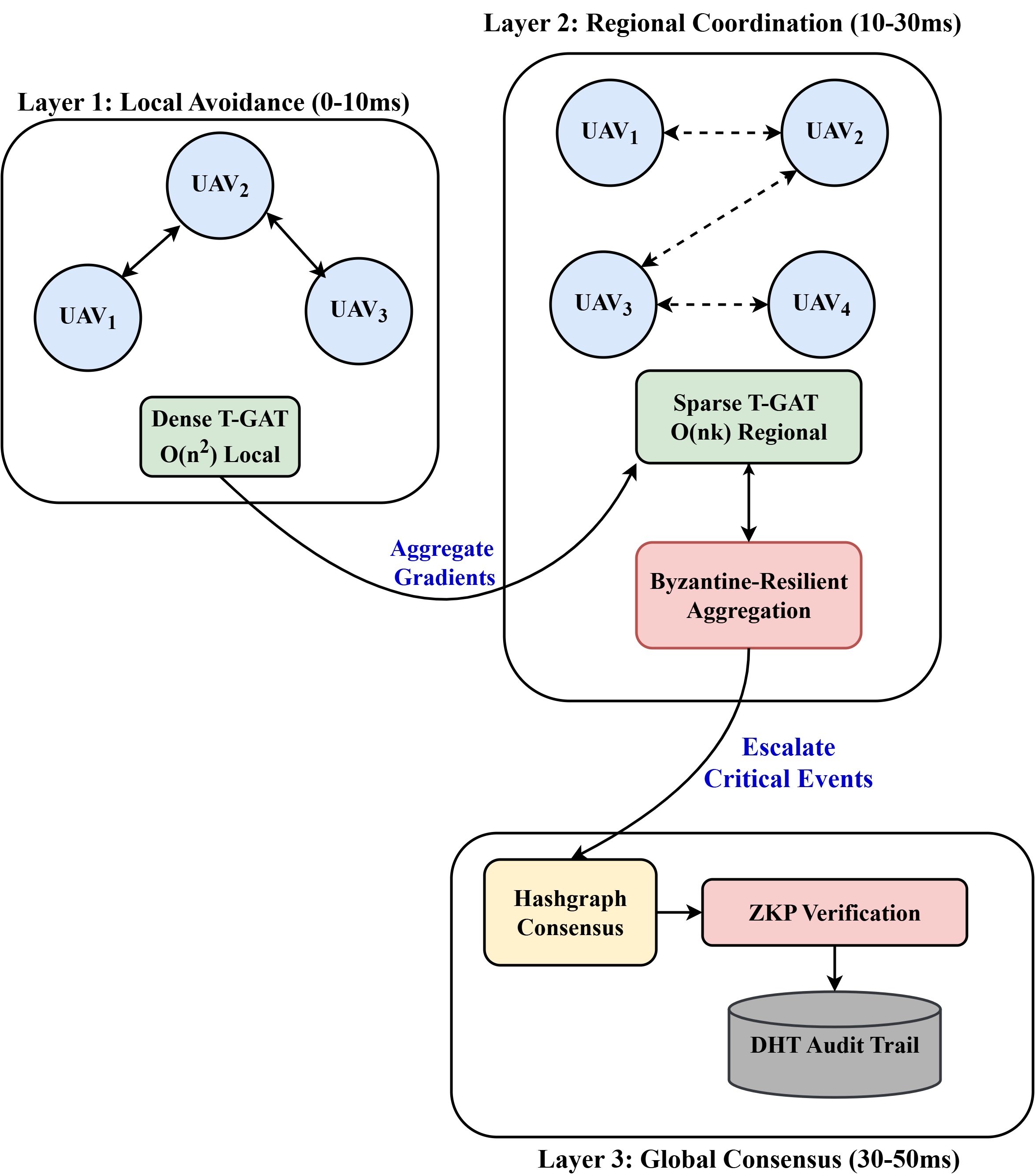}
\caption{Complexity Analysis and Hierarchical Architecture. The three levels of decision-making are nested in each other at escalating scopes of operation and security assurances. Layer 1 adopts the mechanism of dense attention $O(n^2)$ on local neighborhoods, Layer 2 adopts the sparse attention $O(nk)$ on a regional coordination, and Layer 3 offers a global consensus with DHT-based audit trails.}
\label{fig:architecture}
\end{figure*}

We suggest a three-tier hierarchical system where computational and security resources are strategically assigned based on the scope of the decision and latency need. The architecture, termed as HF-GAT (Hierarchical Federated Graph Attention Network), is shown in Fig. \ref{fig:architecture}. This three-layered schema makes it efficiently scalable to distribute the collision avoidance tasks in temporal and spatial dimensions. 

\subsection{Layer 1: Local Collision Avoidance (0-10ms)}
\label{sec21}
The first layer ensures direct peer-to-peer collision avoidance of nearest neighbors, usually in a 5-8 UAVs per 100-meter radius. This layer is the most important in terms of ultra-low latency to a threat of collision.

\subsubsection{\textbf{Dense Temporal Graph Attention}}
\label{sec211}
Each UAV is understood to have its own temporal graph, i.e. $G_1 = (V_1, E_1, T)$, where, $V_1$ is the local peers set of UAVs, $E_1$ is the spatial relationship of the local peers and finally, $T$ is the set of temporal dependencies across observation windows. The attention model operates on sequences of position and velocities $(p_{i,t}, v_{i,t})_{t=1}^T$ that can calculate probabilities of collision of all UAVs. The attention weights are calculated as per Eq. \eqref{eq1}.

\begin{equation}\label{eq1}
\alpha_{ij}^{(t)} = \text{softmax}\left(\frac{(W_q h_i^{(t)})^T (W_k h_j^{(t)})}{\sqrt{d_k}}\right)
\end{equation}
Here, $\alpha_{ij}^{(t)}$ is the attention weight that shows probability of collision between UAVs $i$ and $j$ at time $t$. The spatiotemporal states of UAVs $i$ and $j$ at time $t$ are represented in vectors $h_i^{(t)}$ and $h_j^{(t)}$, respectively. $W_q$ and $W_k$ are the learned query and key transformation matrices, whereas $d_k$ signifies the dimensionality of key vectors participating in the attention scaling.

\subsubsection{\textbf{Security Model}} \label{sec212}
To avoid accurate inference of trajectory with an effective collision avoiding effect, position sharing is subjected to the preservation of basic differential privacy. The privacy parameter is specified as $\epsilon = 0.5$ that regulates the trade-off between the privacy protection and data utility.

\subsection{Layer 2: Regional Coordination (10-30ms)}
\label{sec22}
The second layer provides coordination in regional clusters of 20-50 UAVs where coherent group behavior is attained by computational efficient sparse attention mechanisms. This layer draws a balance between the coordination needs and the computing constraints.

Full attention mechanism has a computational complexity of $O(n^2)$ having $n$ number of UAVs in a cluster. We use a sparse attention operation to reduce the complexity. Each UAV only serves strategically chosen neighbors $k$, computed as $k = \lceil\log_2(n)\rceil$. The neighbor selection strategy incorporates:
\begin{itemize}
\item Geometric distance (3 nearest neighbors according to Euclidean distance)
\item Velocity alignment (2 UAVs with similar vectors of trajectory)
\item Cluster leadership (selected regional coordinators)
\end{itemize}

This sparse attention mechanism restricts the computation complexity within $O(nk)$, while retaining the necessary coordination information required to form swarm coherent behavior.

\subsubsection{\textbf{Asynchronous Federated Learning}} \label{sec221}
We employ an asynchronous Federated Learning (FL) approach to update the global model parameter $\theta$ by the regional servers whenever any participant UAV $i$ uploads its local gradient ($\nabla\theta_i$). Each UAV can update the information without waiting for other UAV participants. The updating rule is given in Eq. \eqref{eq2}.
\begin{equation}\label{eq2}
\theta_{t+1} = \theta_t - \eta \cdot \text{TrimmedMean}(\{\nabla\theta_i\}_{i \in \mathcal{C}_t})
\end{equation}
with $\theta$ being the global model parameter, $\eta$ the learning rate and $\mathcal{C}_t$ the group of UAVs involved at any time $t$. We used a specialized function, known as \emph{TrimmedMean()} where the highest and the lowest 20 percentile coordinate-wise gradient values are dropped to make Byzantine resilient against malicious players.

\subsection{Layer 3: Global Consensus (30-50ms)} \label{sec23}
The third level deals with the global-level planning in the swarm and provides full audit history to check the compliance. Non-time-critical decisions that influence the swarm integrity are handled by this layer.

\subsubsection{\textbf{Hashgraph-Inspired Consensus}} \label{sec231}
When it comes to critical decisions that must be made collectively for all swarm members, a fast Byzantine consensus protocol is used to reach agreement within a 30-50ms latency range. This protocol employs effective gossip communication patterns and virtual voting solutions to achieve consensus despite Byzantine failures.

\subsubsection{\textbf{Distributed Hash Table-Based Audit System}} \label{sec232}
The Zero-Knowledge Proofs (ZKPs) that log the critical actions are stored on a Kademlia Distributed Hash Table (DHT). The generation of ZKP ($\text{proof}$) is formalized in Eq. \eqref{eq3}.
\begin{equation}\label{eq3}
\text{proof}_i = \text{ZKP}(\text{action}_i, \text{state}_i, \text{compliance\_rule})
\end{equation}
where, $\text{action}_i$ is the particular UAV action $i$ under verification, $\text{state}_i$ is the system state during action execution and $\text{compliance\_rule}$ is the regulatory or operational constraint being demonstrated. 

This method offers tamper-resistant audit trails, offering an estimated 85\% reduction in computation and communication overhead as compared to conventional blockchain-based strategies.

\section{Methodology} \label{sec3}
The developed HF-GAT framework brings together a number of major methodological constituents to achieve scalable and resilient UAV collision avoidance. This section discusses the model's working strategies, starting with adaptive privacy mechanism, followed by Byzantine resilient aggregation, and hierarchical decision-making process.

\subsection{Adaptive Differential Privacy Mechanism} \label{sec31}
Generally, fixed noise levels are used as standard procedure in differential privacy, irrespective of the situation of the operation \cite{alfarsi2025}. This is not optimal in the context of the privacy-utility trade-off. The adaptive mechanism proposed here dynamically allocates privacy budgets according to the evaluation that is done in real-time to achieve optimal system performance under different conditions of security.

Each cluster in each region keeps a normalized threat score $\theta_{\text{threat}} \in [0,1]$ with 0 denoting the absence of a threat and 1 indicating an absolute threat. The threat score is calculated as a weighted average of three indicators of the security as given in Eq. \eqref{eq4}
\begin{equation}\label{eq4}
\theta_{\text{threat}} = \alpha \cdot r_{\text{reject}} + \beta \cdot d_{\text{anomaly}} + \gamma \cdot c_{\text{comm}}
\end{equation}
where, $r_{\text{reject}} \in [0,1]$ is the rejection rate of the gradient by Byzantine detection, $d_{\text{anomaly}} \in [0,1]$ anomaly is the deviation of behavior between the expected pattern of UAVs, and $c_{\text{comm}} \in [0,1]$ is a measurement of the communication anomalies such as a delay during message transmission or packet loss. The weighting coefficients $\alpha$, $\beta$, and $\gamma$ are non-negative constants, with $\alpha + \beta + \gamma = 1$. The conventional benchmark is $\alpha = 0.5$, $\beta = 0.3$, and $\gamma = 0.2$ depending on empirical analysis.

Further, an Adaptive privacy budget $\epsilon_{\text{adaptive}}$ has been adopted to increase the privacy protection, as mentioned in Eq. \eqref{eq5}.
\begin{equation}\label{eq5}
\epsilon_{\text{adaptive}} = \epsilon_{\text{min}} + (1 - \theta_{\text{threat}}) \cdot (\epsilon_{\text{max}} - \epsilon_{\text{min}})
\end{equation}
Here, $\epsilon_{\text{min}} = 0.1$ is the smallest value to allow significant privacy protection during maximum threats. $\epsilon_{\text{max}} = 1.0$ is the highest value to allow minimum privacy cost when there are no threats to recognize. 

Such linear interpolation has the advantage of increased privacy protection at both lower and higher threat levels without compromising the decline in the model utility.

\subsection{Asynchronous Byzantine-Resilient Aggregation Protocol} \label{sec32}
The HF-GAT framework uses an asynchronous coordinate-wise \emph{TrimmedMean} aggregation strategy at its regional coordination layer (Layer 2) to be robust to adversarial members. This approach has theoretical immunity to Byzantine failures as well as appropriate computational time that can be used in real-time, and can also adapt to the variation of network latency.

Suppose, a collection of gradient vectors $\{\nabla_1, \nabla_2, \ldots, \nabla_n\}$ is present, where each vector includes the result of a set of participating UAVs $n$. The Byzantine-resilient aggregation is done coordinate-wisely, which means, assign $\{x_{1,j}, x_{2,j}, \ldots, x_{n,j}\}$ (the components of all vectors of the gradient) to each coordinate $j$. The coordinate's $\text{TrimmedMean}_j$ is calculated as Eq. \eqref{eq6}.
\begin{equation}\label{eq6}
\text{TrimmedMean}_j(\{x_{i,j}\}) = \frac{1}{n-2f} \sum_{i=f+1}^{n-f} x_{(i,j)}
\end{equation}
where, $x_{(i,j)}$ is the $i^{th}$ order statistic of the $j^{th}$ coordinate values placed in ascending order, and $f < n/3$ is the maximum possible Byzantine failures that may take place. 

By doing so, this method eliminates the smallest and greatest $f$ values along each coordinate and essentially cleans up any adversarial contribution. Thus, it helps in keeping the central tendency of the otherwise honest contributors. The theory behind this approach is based on the assumption that there are at least $2f+1$ participants out of the total $n$ participants that are honest and agrees with the practical Byzantine Fault Tolerance (pBFT) whereby fewer than 1/3 of nodes should be malicious.

Algorithm \ref{algo1} presents the entire aggregation process that is carried out by the regional servers.
\begin{algorithm}
\caption{Byzantine-Resilient Asynchronous Aggregation}
\label{algo1}
\begin{algorithmic}[1]
\INPUT{Regional server with global model: $\theta_G$}
\OUTPUT{Updated global model: $\theta_G'$}

\Statex \textbf{Initialize:} Global model: $\theta_G$, Byzantine threshold: $f = \lfloor n/3 \rfloor$, Gradient buffer: $\mathcal{B} \leftarrow \emptyset$, learning rate: $\eta$, $\tau_{\text{anomaly}}$ = 0.3
\While{$\text{system\_active}$}
    \State Wait for gradient $\tilde{\nabla}_i$ from UAV $i$
    \State $\text{anomaly\_score} \leftarrow \text{detectAnomaly}(\tilde{\nabla}_i)$
    \If{$\text{anomaly\_score} < \tau_{\text{anomaly}}$}
        \State Add $\tilde{\nabla}_i$ to buffer $\mathcal{B}$
        \If{$|\mathcal{B}| \geq 2f + 1$}
            \State $\bar{\nabla} \leftarrow \text{TrimmedMean}(\mathcal{B})$ \Comment{using Eq. \eqref{eq6}}
            \State $\theta_G \leftarrow \theta_G - \eta \cdot \bar{\nabla}$
            \State Clear buffer $\mathcal{B} \leftarrow \emptyset$
            \State Broadcast updated model $\theta_G$ to regional UAVs
        \EndIf
    \Else
        \State $\theta_{\text{threat}} \leftarrow \text{updateThreatScore}(i)$
        \State Log potential Byzantine behavior from UAV $i$
    \EndIf
\EndWhile
\end{algorithmic}
\end{algorithm}

The protocol uses a gradient buffer $\mathcal{B}$ to receive the gradients $\tilde{\nabla}_i$ sent by the participating UAVs $i$. The gradients with a score $< \text{threshold}\,(\tau_{\text{anomaly}} = 0.3)$ are accepted into the buffer, whereas the other gradients, labeled as suspicious, update the threat score and logging steps.

The Aggregation procedure is triggered as soon as the buffer has at least $2f + 1$ gradients to ensure sufficiency in withstanding the Byzantine failures. The \emph{TrimmedMean} function eliminates the extreme values coordinate-wisely and the following aggregate gradient $\bar{\nabla}$ updates the global model through the learning rate $\eta$. The new model is then transmitted to every regional UAV and the federated learning cycle is complete. Thus, this asynchronism avoids the blocking effect of slow or failed UAV and provides the theoretical Byzantine fault tolerant properties.

\subsection{Hierarchical Decision-Making Process} \label{sec33}
The hierarchical architecture defines the decision-making functions over three different layers, each one with optimal latency and operational scale. Algorithm \ref{algo2} describes the entire chain of hierarchical risk assessment and decision-making process.

\begin{algorithm}
\caption{Hierarchical Risk Assessment and Decision Making}
\label{algo2}
\begin{algorithmic}[1]
\INPUT{UAV state vector: $s_i$, current threat score: $\theta_{\text{threat}}$}
\OUTPUT{Collision avoidance action: $a_i$}
\phase{Local Avoidance (0-10ms)}
\State $\mathcal{N}_{\text{local}} \leftarrow \text{getNearestNeighbors}(s_i, \text{radius})$ \Comment{radius = 100m}
\State $\text{risk}_{\text{local}} \leftarrow \text{DenseT-GAT}(s_i, \mathcal{N}_{\text{local}})$
\If{$\text{risk}_{\text{local}} > \tau_{\text{critical}}$} \Comment{$\tau_{\text{critical}}$=0.8}
    \State \textbf{return} $\text{reflexiveManeuver}(\text{risk}_{\text{local}})$
\EndIf
\phase{Regional Coordination (10-30ms)}
\State $\mathcal{N}_{\text{regional}} \leftarrow \text{getSparseNeighbors}(s_i, k)$ \Comment{$k=\lceil\log_2(n)\rceil$}
\State $\text{context} \leftarrow \text{SparseT-GAT}(s_i, \mathcal{N}_{\text{regional}})$
\State $\nabla_i \leftarrow \text{computeGradient}(\text{experience}_i)$
\State $\epsilon \leftarrow \epsilon_{\text{min}} + (1-\theta_{\text{threat}}) \cdot (\epsilon_{\text{max}} - \epsilon_{\text{min}})$
\State $\tilde{\nabla}_i \leftarrow \text{addDPNoise}(\nabla_i, \epsilon)$
\State $\text{sendAsyncUpdate}(\tilde{\nabla}_i)$
\phase{Global Consensus (30-50ms)}
\If{$\text{criticalEvent}(s_i)$}
    \State $\text{proof} \leftarrow \text{generateZKProof}(a_i, s_i)$
    \State $\text{DHT.store}(\text{hash}(\text{proof}), \text{proof})$
\EndIf
\State \textbf{return} $a_i$
\end{algorithmic}
\end{algorithm}

The Layer 1 implements in-time collision avoidance with dense Temporal Graph Attention Network (T-GAT) in the nearest neighbors within 100-meter radius. The T-GAT receives the current UAV state $s_i$ and local neighborhood information $\mathcal{N}_{\text{local}}$ to compute a local risk score ($risk_{local}$). When it exceeds the critical score $\tau_{\text{critical}}$ (whose value is 0.8), the system is alerted with an instant reflexive maneuver to prevent an upcoming collision.

The Layer 2 organizes the planning at the regional level with sparse attention mechanisms with logarithmic scaling with swarm size. The sparse T-GAT operates on a subsampled set of $k = \lceil\log_2(n)\rceil$  strategically down-sampled neighbors to keep the computation within an acceptable range while retaining critical information about coordination. Here, the system calculates the local gradients according to recent experience as well as adds the noise of adaptive differential privacy. Later, the gradients become a part of the asynchronous federated learning process, as discussed in Section \ref{sec221}.

The Layer 3 manages the global consensus while maintaining the non-time-critical audit trails. Zero-knowledge proofs of critical events are generated and then stored in the distributed hash table (DHT) to be used later in verifications without impairing the privacy of operations.

\section{Experimental Evaluation and Results} \label{sec4}
The experimental verification was made with integrated AirSim and a realistic UAV dynamics model that includes physical-based rotor dynamics, limited communications capabilities within 1km radius, packet losses, and realistic computational constraints on NVIDIA Jetson Xavier NX hardware (8GB RAM, 21 TOPS). Three baseline algorithms were further used for comparative purpose.
\begin{itemize}
\item Centralized DRL (Deep Reinforcement Learning with global state), 
\item Vanilla FedAvg (Standard Federated Learning), and \item Monolithic FGAT (Non-hierarchical Federated GAT with blockchain logging)
\end{itemize}
Three performance metrics - collision rate, mission success rate, and decision latency across swarm sizes between 10 and 500 UAVs in $1km^3$ urban environment were used.

\subsection{Decision Latency Analysis}\label{sec41}
\begin{figure}[!ht]
\centering
\includegraphics[width=0.45\textwidth]{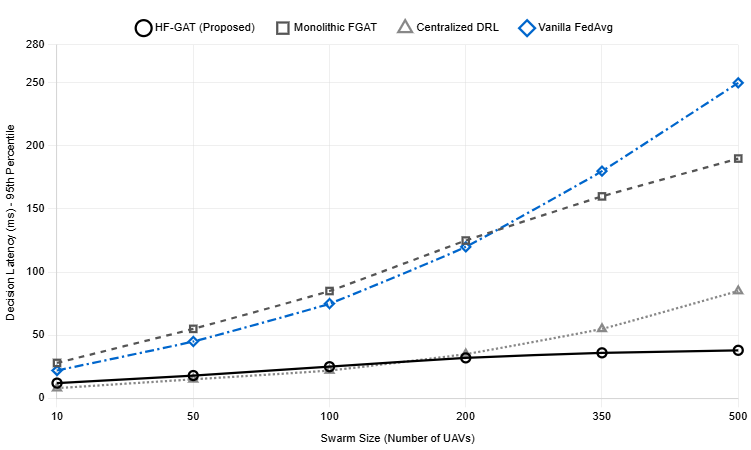}
\caption{Decision latency comparison across different swarm sizes}
\label{fig:r1}
\end{figure}

Figure \ref{fig:r1} presents the decision latency (measured in milliseconds) across different swarm sizes. The proposed HF-GAT framework maintains 95th percentile decision latency (below 40ms) with a swarm of 500 UAVs. This is approximately 70\% lower than Monolithic FGAT. The hierarchical design of HF-GAT only confines dense computations to the local neighborhood and make use of sparse attention to coordinate regions, thereby achieving up to 92\% reductions in communication overheads in large swarms. 

Despite having the lowest latency in an optimal scenario, Centralized DRL suffers latent communication requirements leading to bottlenecks in addition to single point of failures. Further, the HF-GAT achieves a consistently sub-50ms decision level on all the scales tested and so has a practical applicability to time-sensitive applications.

\subsection{Adversarial Byzantine Resilience Analysis} \label{sec42}
\begin{figure}[!ht]
\centering
\includegraphics[width=0.45\textwidth]{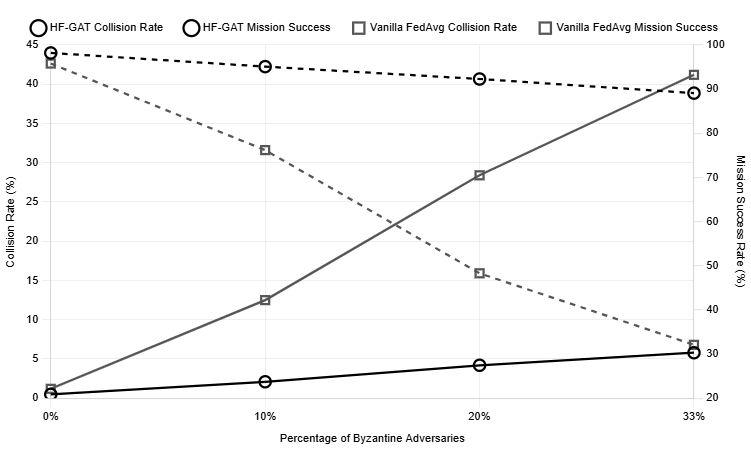}
\caption{Performance under Byzantine adversarial attacks}
\label{fig:r2}
\end{figure}

Figure \ref{fig:r2} shows the resilience to Byzantine attacks in terms of collision rate (measured in \%). The adversarial testing substantiates the survival of the proposed HF-GAT model against the threats. The nodes are using gradient poisoning and false position reporting to simulate a Byzantine attack. HF-GAT ensures that the collision rates do not exceed 6\% with 89.1\% mission success rate. This vastly exceeds the catastrophic failure of Vanilla FedAvg having collision rate of over 40\%. 

The high survivability of HF-GAT lies in the incorporation of Byzantine-resilient aggregate at the regional layer and architectural isolation that narrows the range of the blast of local adversaries.The asynchronous nature of HF-GAT complicates the predictability of attack windows and continues to operate effectively.

\subsection{Success Rate Analysis} \label{sec43}
\begin{figure}[!ht]
\centering
\includegraphics[width=0.45\textwidth]{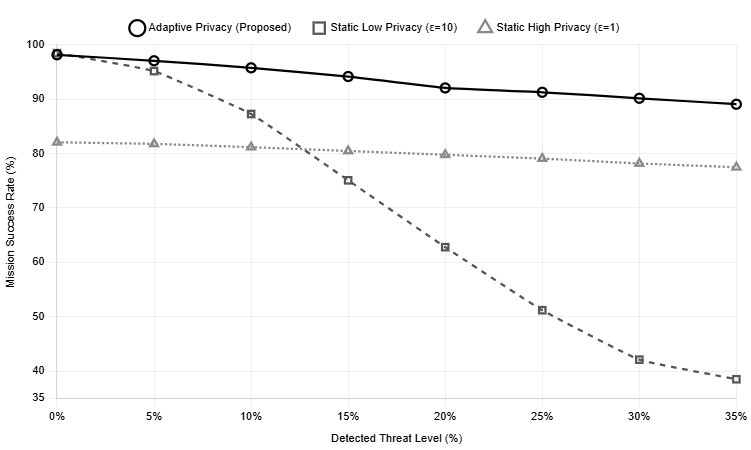}
\caption{Adaptive privacy mechanism performance under varying threat levels}
\label{fig:r3}
\end{figure}

Adaptive privacy mechanism performs effective optimization of the privacy-utility trade-off, which can be seen in Fig. \ref{fig:r3}. On a low-threat regime, the adaptive strategy of HF-GAT model has a mission success rate almost equal to the static low-privacy arrangements with added security defense available in the event of a threat's detection. Further, an average of 12-18\% increased model accuracy over the static high-privacy alternatives can also be devised from the figure. 

This substantiates our claim that the proposed HF-GAT schema can dynamically specify privacy budgets according to real-time threats. This flexible feature would help the privacy defense to adapt to the different threat scales such that it would not cause excessive performance compromise in normal operations, yet has a high level security assurance when attacked.

\section{Conclusion}\label{sec5}
This paper discussed the trilemma of scalability, security and UAV collision avoidance by designing a three-layered hierarchical architecture. Our construction has sub-linear scaling $O(nk)$ due to the use of sparse attention, Byzantine fault tolerance with resilience to $f<n/3$ adversaries, and adaptive privacy preservation. Experiment shows a 70\% decrease in latency when compared to the monolithic techniques with collision rates of $< 2\%$ under normal conditions and $< 6\%$ during worst cases. 

The hierarchical strategy of the proposed HF-GAT model allows deploying 500-UAV swarms in practice with real-time performance, which in turn validate our claim of a scalable autonomous system. Future research will be performed on hardware-in-the-loop validation, safety verification and energy-aware resource allocation to long missions.



\begin{thebibliography}{10}
	\providecommand{\url}[1]{#1}
	\csname url@samestyle\endcsname
	\providecommand{\newblock}{\relax}
	\providecommand{\bibinfo}[2]{#2}
	\providecommand{\BIBentrySTDinterwordspacing}{\spaceskip=0pt\relax}
	\providecommand{\BIBentryALTinterwordstretchfactor}{4}
	\providecommand{\BIBentryALTinterwordspacing}{\spaceskip=\fontdimen2\font plus
		\BIBentryALTinterwordstretchfactor\fontdimen3\font minus
		\fontdimen4\font\relax}
	\providecommand{\BIBforeignlanguage}[2]{{%
			\expandafter\ifx\csname l@#1\endcsname\relax
			\typeout{** WARNING: IEEEtran.bst: No hyphenation pattern has been}%
			\typeout{** loaded for the language `#1'. Using the pattern for}%
			\typeout{** the default language instead.}%
			\else
			\language=\csname l@#1\endcsname
			\fi
			#2}}
	\providecommand{\BIBdecl}{\relax}
	\BIBdecl
	
	\bibitem{chang2023distributed}
	H.~Chang, Y.~Liu, and Z.~Sheng, ``Distributed multi-agent reinforcement
	learning for collaborative path planning and scheduling in blockchain-based
	cognitive internet of vehicles,'' \emph{IEEE Transactions on Vehicular
		Technology}, vol.~73, no.~5, pp. 6301--6317, 2024.
	
	\bibitem{talaat2024blockchain}
	F.~M. Talaat and A.~A. Hamza, ``Blockchain-enhanced artificial intelligence for
	advanced collision avoidance in the internet of vehicles (iov),''
	\emph{Neural Computing and Applications}, vol.~37, no.~6, pp. 4915--4936,
	2025.
	
	\bibitem{zeng2025intersection}
	M.~Zeng, M.~S.~M. Mohamad~Hashim, M.~N. Ayob, A.~H. Ismail, and Q.~Zang,
	``Intersection collision prediction and prevention based on
	vehicle-to-vehicle (v2v) and cloud computing communication,'' \emph{PeerJ
		Computer Science}, vol.~11, p. e2846, 2025.
	
	\bibitem{guo2025hybrid}
	C.~Guo, J.~Wu, P.~Luo, Z.~Wang, K.~Zhang, Z.~Yang, Z.~Dou, and K.~Song,
	``Hybrid reinforcement learning-based collision avoidance algorithm for
	autonomous vehicle clusters,'' \emph{IEEE Access}, vol.~13, pp.
	61\,564--61\,577, 2025.
	
	\bibitem{alagha2025uav}
	A.~Alagha, M.~Kadadha, R.~Mizouni, S.~Singh, J.~Bentahar, and H.~Otrok,
	``Uav-assisted internet of vehicles: A framework empowered by reinforcement
	learning and blockchain,'' \emph{Vehicular Communications}, vol.~52, p.
	100874, 2025.
	
	\bibitem{sunshine2019rvo}
	B.~Sunshine-Hill, ``Rvo and orca: How they really work,'' in \emph{Game AI Pro
		360: Guide to Movement and Pathfinding}.\hskip 1em plus 0.5em minus
	0.4em\relax CRC Press, 2019, pp. 245--256.
	
	\bibitem{wang2025blockchain}
	X.~Wang, H.~Zhu, Z.~Ning, L.~Guo, and Y.~Zhang, ``Blockchain intelligence for
	internet of vehicles: Challenges and solutions,'' \emph{IEEE Communications
		Surveys \& Tutorials}, vol.~25, no.~4, pp. 2325--2355, 2023.
	
	\bibitem{guillen2022multi}
	A.~Guillen-Perez and M.-D. Cano, ``Multi-agent deep reinforcement learning to
	manage connected autonomous vehicles at tomorrows intersections,'' \emph{IEEE
		Transactions on Vehicular Technology}, vol.~71, no.~7, pp. 7033--7043, 2022.
	
	\bibitem{kwon2019multi}
	D.~Kwon and J.~Kim, ``Multi-agent deep reinforcement learning for cooperative
	connected vehicles,'' in \emph{2019 IEEE Global Communications Conference
		(GLOBECOM)}, 2019, pp. 1--6.
	
	\bibitem{lu2020blockchain}
	Y.~Lu, X.~Huang, K.~Zhang, S.~Maharjan, and Y.~Zhang, ``Blockchain empowered
	asynchronous federated learning for secure data sharing in internet of
	vehicles,'' \emph{IEEE Transactions on Vehicular Technology}, vol.~69, no.~4,
	pp. 4298--4311, 2020.
	
	\bibitem{li2021byzantine}
	Z.~Li, H.~Yu, T.~Zhou, L.~Luo, M.~Fan, Z.~Xu, and G.~Sun, ``Byzantine resistant
	secure blockchained federated learning at the edge,'' \emph{IEEE Network},
	vol.~35, no.~4, pp. 295--301, 2021.
	
	\bibitem{liu2020blockchain}
	Y.~Liu, F.~R. Yu, X.~Li, H.~Ji, and V.~C.~M. Leung, ``Blockchain and machine
	learning for communications and networking systems,'' \emph{IEEE
		Communications Surveys \& Tutorials}, vol.~22, no.~2, pp. 1392--1431, 2020.
	
	\bibitem{wang2025fedfr}
	D.~Wang and S.~Guan, ``Fedfr-adp: Adaptive differential privacy with feedback
	regulation for robust model performance in federated learning,''
	\emph{Information Fusion}, vol. 116, p. 102796, 2025.
	
	\bibitem{alfarsi2025}
	A.~S.~S. Al~Farsi, A.~Khan, M.~R. Mughal, and M.~M. Bait-Suwailam, ``Privacy
	and security challenges in federated learning for uav systems: A systematic
	review,'' \emph{IEEE Access}, vol.~13, pp. 86\,599--86\,615, 2025.
	
\end{thebibliography}
\end{document}